\documentclass[10pt,twocolumn,letterpaper]{article}

\usepackage{cvpr}              
\usepackage{tabu}                     
\usepackage{multirow}                 
\usepackage{multicol}                 
\usepackage{multirow}                
\usepackage{float}                    
\usepackage{makecell}                 
\usepackage{booktabs}                 
\usepackage{color}
\usepackage{bbding}
\usepackage[dvipsnames]{xcolor}

\definecolor{cvprblue}{rgb}{0.21,0.49,0.74}
\usepackage[pagebackref,breaklinks,colorlinks,citecolor=cvprblue]{hyperref}

\title{Learning to See Low-Light Images via Feature Domain Adaptation}
\author{Qirui Yang\\
Tianjin University\\
\and
Qihua Cheng\\
Shenzhen MicroBT Electronics Technology Co., Ltd\\
\and
Huanjing Yue\\
Tianjin University\\
\and
Le Zhang\\
Shenzhen MicroBT Electronics Technology Co., Ltd\\
\and
Yihao Liu\\
Shanghai Artificial Intelligence Laboratory\\
\and
Jingyu Yang\\
Tianjin University\\
}

\begin{document}
\maketitle

\begin{abstract}
Raw low light image enhancement (LLIE) has achieved much better performance than the sRGB domain enhancement methods due to the merits of raw data. However, the ambiguity between noisy to clean and raw to sRGB mappings may mislead the single-stage enhancement networks. The two-stage networks avoid ambiguity by decoupling the two mappings but usually have large computing complexity. To solve this problem, we propose a single-stage network empowered by Feature Domain Adaptation (FDA) to decouple the denoising and color mapping tasks in raw LLIE. The denoising encoder is supervised by the clean raw image, and then the denoised features are adapted for the color mapping task by an FDA module. We propose a Lineformer to serve as the FDA, which can well explore the global and local correlations with fewer line buffers (friendly to the line-based imaging process). During inference, the raw supervision branch is removed. In this way, our network combines the advantage of a two-stage enhancement process with the efficiency of single-stage inference. 
Experiments on four benchmark datasets demonstrate that our method achieves state-of-the-art performance with fewer computing costs (60\% FLOPs of the two-stage method DNF). \textit{Our codes will be released after the acceptance of this work.}
\end{abstract}
    
\section{Introduction}
\label{sec:intro}

Low-light image enhancement (LLIE) is a pervasive but challenging problem that aims to improve the visibility of low-light images and reveal the details obscured by the darkness. However, imaging systems encounter challenges in low-light environments due to low photoelectric conversion efficiency and complicated noise. Although increasing exposure time can improve the signal-to-noise ratio of low-light images, it may lead to unwanted motion blur.

\begin{figure}[t]
    \centering	
    \centering{\includegraphics[width=8.35cm]{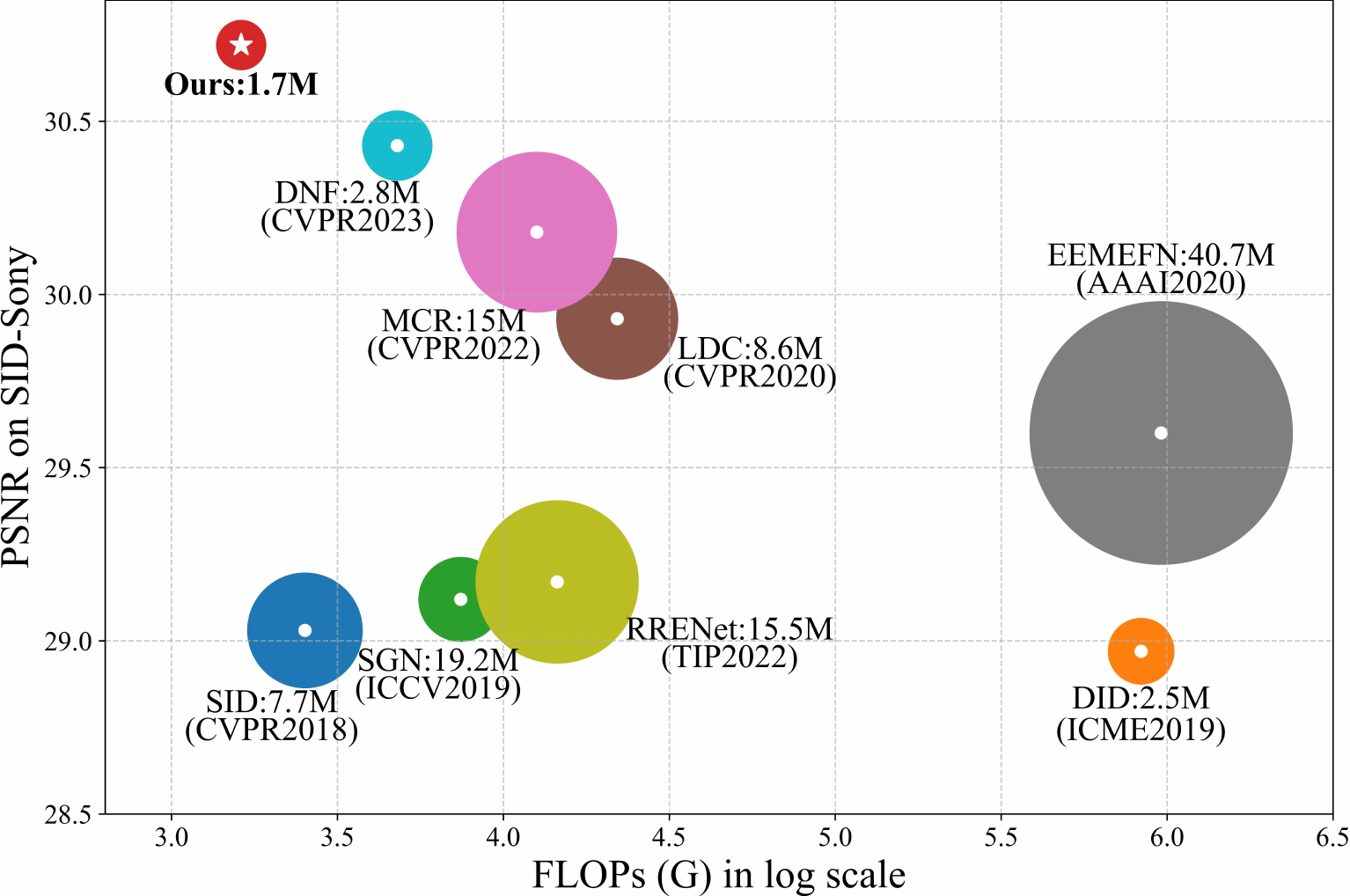}}
    \caption {Comparison of PSNR, FLOPs, and Parameters by different methods on the SID-Sony dataset.}
    \label{feature}
\end{figure}

During the past decade, the LLIE paradigm has shifted from traditional optimization-based enhancement methods \cite{park2017low, Wei_Wang_Yang_Liu_2018} to learning-based methods \cite{chen2018learning, xu2020learning, maharjan2019improving}, and the enhancement performance has been greatly improved. Since the low-light images are heavily degraded by noise, the enhancement process usually involves two tasks, i.e., noise removal and color intensity restoration. Considering that the noise distribution in raw images is much simpler than that in the sRGB domain and the raw images have larger bit depth, the raw-based image enhancement has garnered increasing attention and yielded promising results
\cite{Huang_Yang_Hu_Liu_Duan_2022, Yue_Cao_Liao_Chu_Yang_2020}. Therefore, in this work, we focus on LLIE with raw inputs.

\begin{figure}[t]
    \centering	   
    \centering{\includegraphics[width=8.5cm]{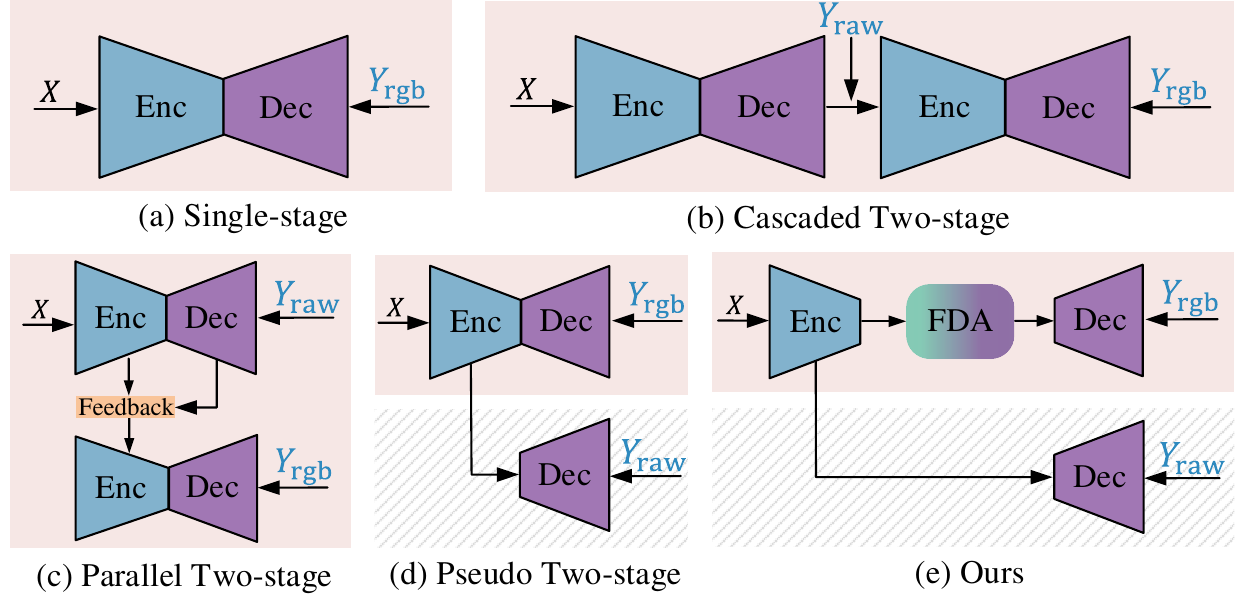}}  
    \caption {Comparison of common paradigms of raw LLIE models. Cascaded, Parallel, and Pseudo two-stage have the same raw supervision. $X$, $Y_{\text{raw}}$, and $Y_{\text{rgb}}$ denote the raw noisy input, raw ground truth and sRGB ground truth, respectively. The pink region indicates the inference pipeline for each method.}
    \label{frame}
\end{figure}

The raw-based image enhancement methods can be classified into three categories: single-stage based, cascaded two-stage based, and parallel two-stage based, as shown in Fig. \ref{frame}. Specifically, the single-stage methods \cite{chen2018learning, maharjan2019improving} directly map noisy raw images to clean sRGB images via an encoder-decoder structure. This kind of framework has fewer parameters and is computationally efficient. However, they often suffer from feature domain misalignment during cross-domain (namely raw to sRGB domain) mapping, leading to subpar performance. To address this issue, two-stage networks \cite{dong2022abandoning, xu2020learning, jin2023dnf} have emerged, which decompose the raw-based LLIE task into two sub-tasks: raw denoising and raw-sRGB color mapping. Among them, the cascaded two-stage networks \cite{Huang_Yang_Hu_Liu_Duan_2022, Liang_Cai_Cao_Zhang_2019} 
utilize the first encoder-decoder to solve the raw denoising task by utilizing the raw ground truth as the supervision, and then utilize the second encoder-decoder to solve the color mapping task by utilizing the final sRGB ground truth as supervision. The two tasks are jointly optimized together and have achieved better results than the single-stage methods. However, this may lead to information loss and cumulative errors since the second stage can only get the image level result of the first stage. In contrast, the parallel two-stage methods \cite{jin2023dnf, dong2022abandoning} utilize two parallel encoder-decoders, where the top encoder-decoder is used for noise removal, and the bottom encoder-decoder is used for denoising and color mapping. Different from the cascaded structure, the inputs of the bottom (second-stage) encoder are the features of the top (first-stage) encoder-decoder other than the result of the first stage to avoid the information loss problem. In addition, the parameters of the two encoders are shared to make the second encoder also have a denoising ability. However, this strategy also has a large computing complexity. 

In summary, the single-stage networks are efficient but cannot satisfactorily perform the color mapping and noise removal tasks simultaneously. These two-stage networks decouple the denoising and color mapping tasks but with large computing complexity. \textbf{A natural question arises: can we utilize one single-stage network to decouple the two tasks?} A naive solution is shown in Fig. \ref{frame} (d), namely utilizing one shared encoder and two separated decoders for color mapping and denoising tasks, respectively. However, the encoder features are still ambiguous since they need to fit two different tasks. To solve this problem, we propose a feature adaptation strategy to bridge the denoising features and color mapping features, namely the Feature Domain Adaptation Network (FDANet), as shown in Fig. \ref{frame} (e). We decouple the denoising and color mapping tasks by utilizing two decoders and one Feature Domain Adaptation module. In this way, the encoder features are first tailored for the denoising task, and after feature adaptation, they are suitable for the color mapping task. Note that, it is not trivial to design the feature adaptation module since it is required to focus on local noise and details and balancing global brightness and color tone, while having less computational costs. In this work, we propose a Line-Based Transformer (termed as Lineformer), which explores the global correlation on the line dimension and the local correlation on the height dimension. In addition, it is friendly to the line-based imaging process due to fewer line buffers. 

Our main contributions are as follows. 
\begin{itemize}
\item[$\bullet$] We propose a novel single-stage network to decouple the denoising and color mapping tasks in raw LLIE, which combines the advantage of a two-stage enhancement process with the efficiency of single-stage inference. 
\item[$\bullet$] We propose an FDA module to bridge denoising and color mapping features. The FDA module is constructed with Lineformer, which explores both global and local correlations with fewer line buffers. 
\item[$\bullet$] Our method outperforms state-of-the-art raw LLIE methods with only 60$\%$ of the parameters and FLOPs of two-stage methods.
\end{itemize}

\section{Related Work}
\label{sec:related}

\subsection{Raw Low-Light Image Enhancement}
Raw LLIE tasks have received increasing attention due to the growing demand for digital camera image processing. Early studies were usually single-stage approaches. Chen et al. \cite{chen2018learning} proposed the first end-to-end raw LLIE method, which captured a large-scale paired dataset. This work inspired several subsequent works that improved SID using different network architectures or loss functions. For instance, SGN \cite{gu2019self} proposed a self-guided neural network that balanced denoising performance and computational cost by utilizing the large-scale contextual information from the shuffled multi-resolution input. DID \cite{maharjan2019improving} replaced the U-Net in SID with residual learning to better preserve the information in image features. Lamba et al. \cite{Lamba_Balaji_Mitra_2020, Lamba_Mitra_2021} focused on developing fast inference frameworks and achieving real-time inference performance.

Although single-stage methods have a simple inference framework, they often perform poorly since raw LLIE involves both image denoising and color mapping tasks. Therefore, two-stage methods were proposed to obtain better performance. LDC \cite{xu2020learning} implemented a frequency-based decomposition strategy that filters out high-frequency features first, recovers low-frequency features with an amplification operation, and then restores high-frequency details. RRENet \cite{Huang_Yang_Hu_Liu_Duan_2022} proposed a raw-guided exposure enhancement network based on the ISP pipeline flow. MCR \cite{dong2022abandoning} generated monochrome images from color raw images for intermediate supervision. DNF \cite{jin2023dnf} introduced a parallel two-stage method to avoid information loss. However, these frameworks usually need two encoder-decoder structures, which is not computationally efficient. 

\subsection{Raw Image Based Restoration}
Due to the merits of raw data, raw-based restoration \cite{liu2023joint,yang2023efficient} has made significant progress in recent years. Similar to raw image enhancement, these tasks involve two separate tasks: raw image restoration \cite{Wei_Fu_Yang_Huang_2020} and raw-sRGB domain mapping \cite{schwartz2018deepisp, Liu_Feng_Wang_2022}. These networks either utilize an end-to-end network to accomplish the two tasks simultaneously \cite{yue2022real,wang2020practical,zhang2019zoom} or utilize cascaded networks \cite{Yue_Cao_Liao_Chu_Yang_2020,liang2020raw}, the first network performs the restoration task and the second network (or traditional module) performs the raw to sRGB domain mapping task. For example, The works in \cite{zhang2019zoom, yue2022real} use an end-to-end network to perform the raw image (video) super-resolution task. The work in \cite{yue2022recaptured} first performs the demoireing task in the raw domain and then utilizes a pretrained ISP module to transform the result into the sRGB domain. However, the cascaded frameworks have large computation costs and introduce information loss problems. In contrast, the end-to-end single-stage networks may degrade the restoration performance since they must perform the two tasks simultaneously. In this paper, we propose to solve the domain gap problem, which is common in raw image processing, from the perspective of domain adaptation. This strategy is not only insightful for raw-based LLIE task, but also for the other raw image processing tasks.
   
\subsection{Domain Adaptation}

Domain adaptation is a well-studied topic, which aims to realize the transformation from the source domain to the target domain to enhance the results on the target task \cite{luo2021category, wu2023dafd}. For example, domain adaptation enables a classification model to work in both source and target domains by modifying the features extracted from the target domain through contrastive learning \cite{kang2019contrastive} or adversarial learning \cite{chen2022reusing, du2020dual, xu2020adversarial}. In this way, the domain gap between the target and source features is reduced. Inspired by this, we propose feature domain adaptation for raw LLIE to bridge the features from the raw encoder to the sRGB decoder. Different from unsupervised domain adaptation, we have labels for the target domain. Therefore, we utilize two parallel supervisions to optimize the domain adaptation module. In addition, we propose a Lineformer to make the FDA module capture both local and global correlations efficiently.

\section{Method}
\label{sec:method}

\subsection{Motivation} 
\label{sec3.1}

As introduced in Sec. \ref{sec:intro}, the single-stage framework is effective but cannot simultaneously perform color mapping and noise removal tasks well. The two-stage frameworks decouple the denoising and color mapping tasks, achieving better performance. To investigate the reasons and challenges behind these phenomena, we analyze the popular paradigm of raw LLIE frameworks by comparing their feature representation differences. 

\begin{figure}[ht]
    \centering	   
    \centering{\includegraphics[width=8.3cm]{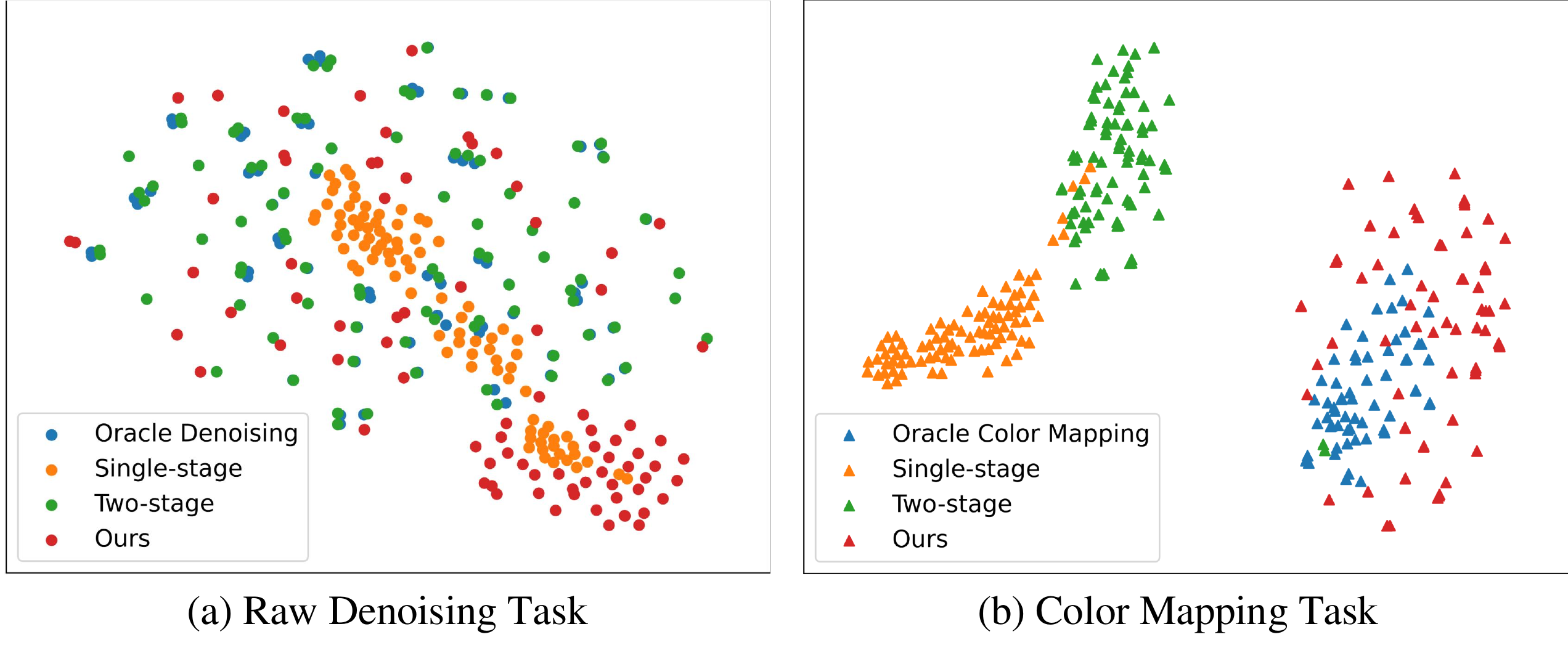}} 
    \caption {Comparison of encoded features. The features $F^{\text{orDe}}_{\text{raw}}$, $F^{1}_{\text{raw}}$, and $F^{\text{2Dn}}_{\text{raw}}$ for the denoising task are denoted as blue, orange, and green points in (a), respectively. The features $F^{\text{orCM}}_{\text{rgb}}$, $F^{1}_{\text{raw}}$, and $F^{\text{2CM}}_{\text{rgb}}$ for the color mapping task are denoted as blue, orange, and green triangles in (b), respectively.}
    \label{sacatter}
\end{figure}

First, we give two oracle experiments as the benchmark. We optimize the denoising and color mapping tasks using two single-stage frameworks. The denoising network maps the noisy raw images to clean raw images. The color mapping network maps the clean raw to clean sRGB images. The results of the two models can serve as the upper bound for the two tasks. Then, for the denoising task, we compare the features extracted from the encoder of the oracle denoising network ($F^{\text{orDe}}_{\text{raw}}$), the encoder of the single-stage enhancement network shown in Fig. \ref{frame} (a) ($F^{1}_{\text{raw}}$), the first encoder of the cascaded two-stage network shown in Fig. \ref{frame} (b) ($F^{\text{2Dn}}_{\text{raw}}$). To ease the comparison, we use t-distributed Stochastic Neighbor Embedding (t-SNE) \cite{Donahue_Jia_2013,liu2021discovering, Wang_Cao_Zha_2020} to visualize these high-dimensional features. 
The results are shown in Fig. \ref{sacatter} (a). It can be observed that $F^{\text{2Dn}}_{\text{raw}}$ is close to the benchmark features $F^{\text{orDe}}_{\text{raw}}$ for the denoising task since $F^{\text{2dn}}_{\text{raw}}$ is also optimized by the denoising task. In contrast, $F^{1}_{\text{raw}}$ is far from $F^{\text{orDe}}_{\text{raw}}$ since it is optimized for both denoising and color mapping tasks. 

\begin{figure}[ht]
    \centering	   
    \centering{\includegraphics[width=8.3cm]{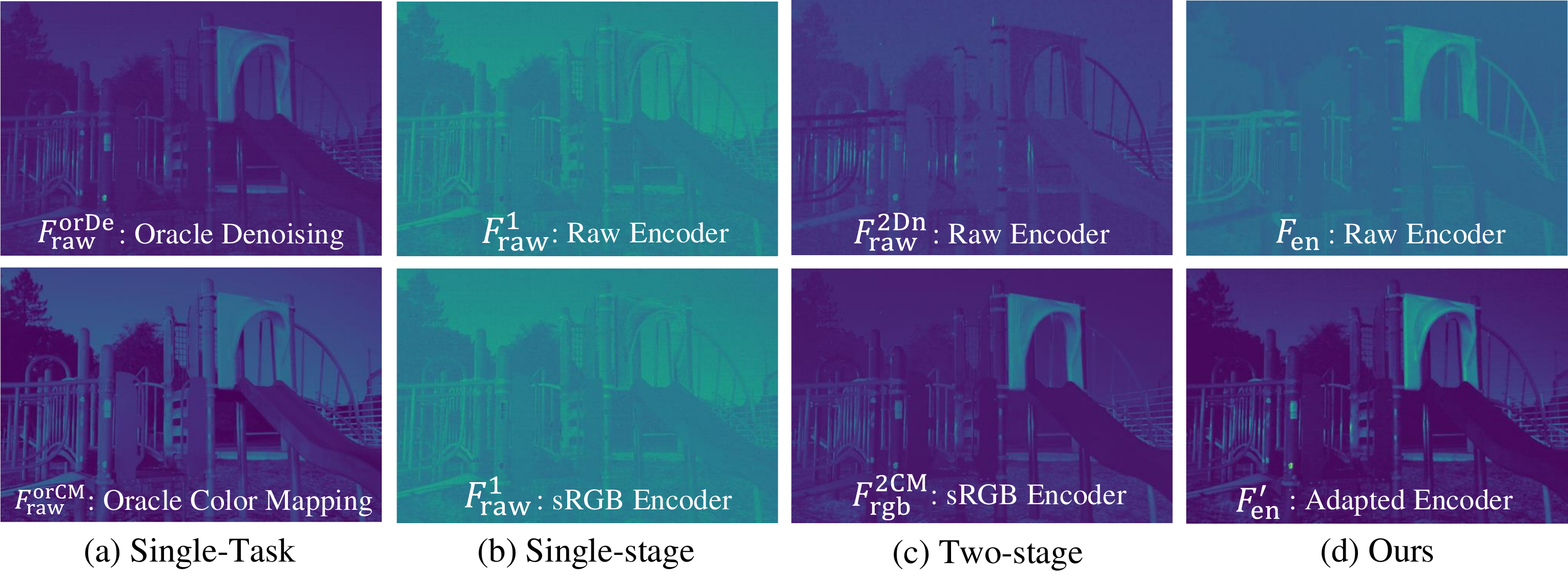}} 
    \caption{Visualization of the features extracted by the corresponding encoder for different frameworks. The symbols denote the same features as those in Fig. \ref{sacatter}. The raw encoder means the input of the encoder is original raw noisy images. The sRGB encoder means the output of the encoder is directly fed into a sRGB decoder. Therefore, the Raw and sRGB encoders are the same for the single-stage method (b).} 
    \label{feature_map}
\end{figure}

\begin{figure*}[t]
    \centering	   
    \centering{\includegraphics[width=17cm]{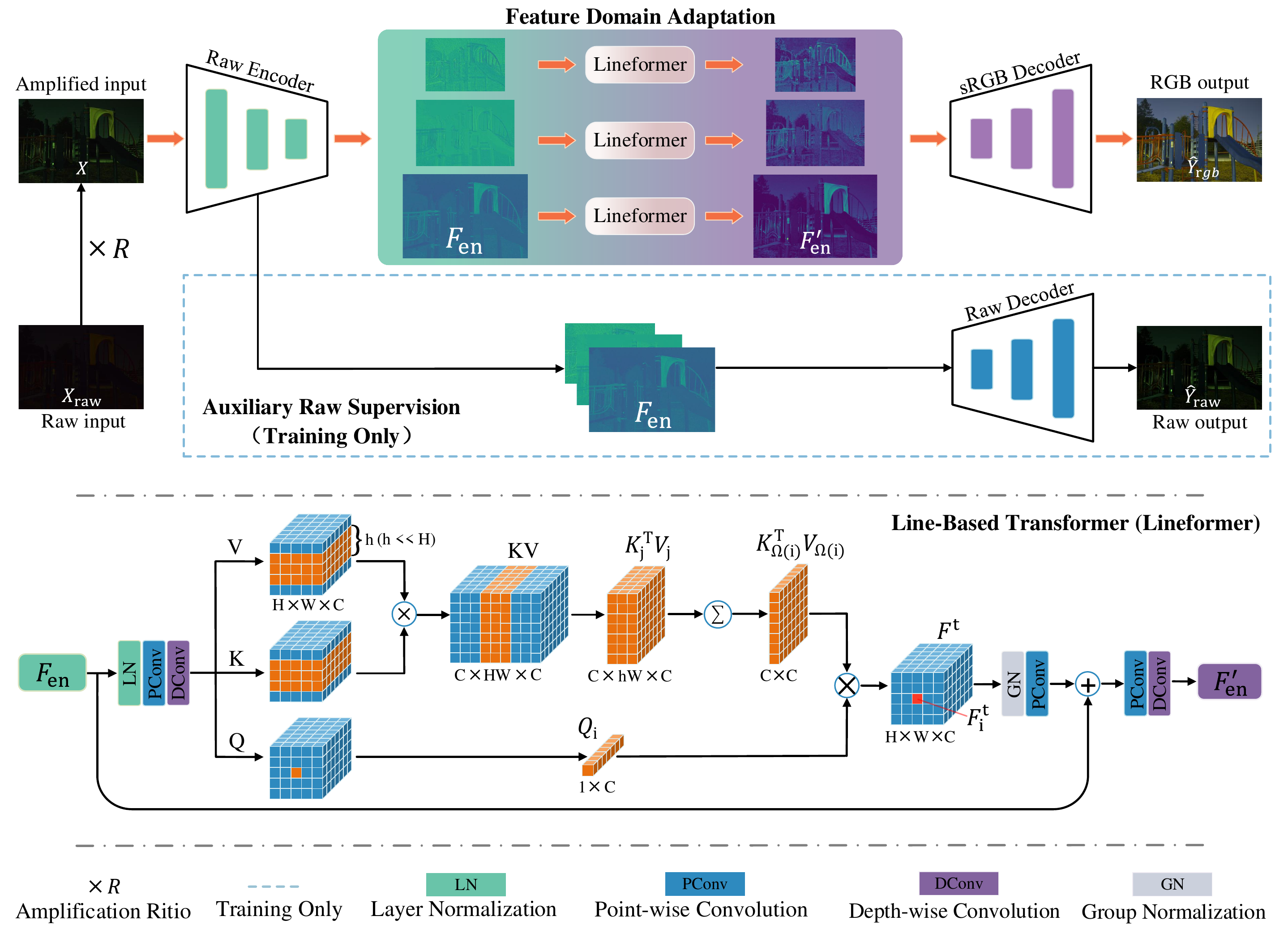}}  
    \caption {Overview of the proposed FDANet. The first part is FDANet, which contains a color mapping branch (top branch) with a Feature Domain Adaptation module and a raw denoising branch (bottom branch). Note that the auxiliary raw supervision network is removed when inference. The second part is the detailed structure for the proposed Lineformer.}
    \label{model}
\end{figure*}

Similarly, for the color mapping task, we compare the features extracted from the encoder of the oracle color mapping network ($F^{\text{orCM}}_{\text{rgb}}$), the encoder of the single-stage enhancement network shown in Fig. \ref{frame} (a) ($F^{1}_{\text{raw}}$), the second encoder of the cascaded two-stage network shown in Fig. \ref{frame} (b) ($F^{\text{2CM}}_{\text{rgb}}$). The results are shown in Fig. \ref{sacatter} (b). It can be observed that both $F^{1}_{\text{raw}}$ and  $F^{\text{2CM}}_{\text{rgb}}$ are all far from $F^{\text{orCM}}_{\text{rgb}}$. This implies that the encoded features from the single-stage and two-stage networks are still far from the upper bound for the color mapping task. 

To solve this problem, we propose a feature domain adaptation strategy to bridge the gap between denoising features and color mapping features. After feature adaptation, our encoding features (domain adapted features) used for color mapping approximate the distribution of the oracle color mapping task. Meanwhile, our denoising features (the features of our encoder) approximate both the oracle denoising features $F^{\text{orDe}}_{\text{raw}}$ and the single-stage features $F^{1}_{\text{raw}}$ since our denoising features are optimized by both raw and sRGB domain supervisions. Fig. \ref{feature_map} further presents visual examples of these features. The feature $F^{\text{orCM}}_{\text{rgb}}$ has the most vivid textures, while the features $F^{1}_{\text{raw}}$ and $F^{\text{2CM}}_{\text{rgb}}$ are smooth. Our feature adaptation module can make the encoded features approximate $F^{\text{orCM}}_{\text{rgb}}$. The details about our solution are given in the following section. 

\subsection{Overview of Our FDANet}
Our FDANet is shown in Fig. \ref{model}, which contains a color mapping branch (top branch) and a denoising branch (bottom branch). The input of the network is the low-light raw image $X_{raw}$ amplified by a predetermined magnification factor $R$, which is denoted as $X$. Then, the raw encoder $E_{\text{raw}}$ encodes $X$ to obtain multi-scale features $F_{\text{en}}$. Since we aim to utilize a single-stage network to accomplish both the denoising and color mapping tasks, the feature $F_{\text{en}}$ should fit the two tasks. Therefore, at the top branch, we propose a Feature Domain Adaptation (FDA) module to map raw encoder adaptively features to the sRGB domain, which overcomes the feature domain misalignment problem. The adapted features $F^{'}_{\text{en}}$ are then fed into the sRGB decoder $D_{\text{rgb}}$ to generate the final output $\hat{Y}_{\text{rgb}}$ in the sRGB domain. In order to remove noise in the encoded features $F_{\text{en}}$, we introduce an auxiliary raw supervision network that feeds the multi-scale features $F_{\text{en}}$ into the auxiliary raw decoder $D_{\text{raw}}$ to output $\hat{Y}_{\text{raw}}$. The two branches are training together. During the inference phase, the auxiliary raw supervision network is deactivated, making our network more efficient than two-stage enhancement methods. In the following, we present details of the two branches of FDANet.  

\begin{table*}[t]
    \centering
    \caption{Quantitative results of raw-based LLIE methods on the SID-Sony \cite{chen2018learning} and LRID \cite{feng2023learnability} datasets. The FLOPs are measured on the whole raw image of the SID dataset with a resolution of $2848 \times 4256$. The best result is in red \textbf{{\color{red}bold}} whereas the second best one is in blue {\color{blue}\underline{underlined}}. Metrics with $\uparrow$ and $\downarrow$ denote higher better and lower better, respectively.}
    \scalebox{0.9}{
        \begin{tabular}{clcccccccccc}
            \hline
            \multirow{2}{*}{Category} & \multirow{2}{*}{Method} & \multirow{2}{*}{FLOPs} & \multirow{2}{*}{\#Param} & \multicolumn{4}{c}{Sony\cite{chen2018learning}}  & \multicolumn{4}{c}{LRID\cite{feng2023learnability}}\\  
            \cline{5-12}
            &     &  &     &  PSNR$\uparrow$  & SSIM$\uparrow$ & $\Delta E$$\downarrow$  & LPIPs$\downarrow$ &  PSNR$\uparrow$  & SSIM$\uparrow$ & $\Delta E$$\downarrow$  & LPIPs$\downarrow$\\  
            \hline      
            \multirow{5}{*}{Single-Stage} 
            & \multicolumn{1}{l}{LLPackNet \cite{Lamba_Balaji_Mitra_2020}}   &83.3G  &1.2M   &28.59	&0.7817	&7.353	&0.1912  &31.60	&0.8696	&4.723	& 0.2927 \\
            & \multicolumn{1}{l}{IRT \cite{Lamba_Mitra_2021}}  &59.6G    & 0.8M    &28.83	&0.7796	&7.158	&0.1777  &31.48	&0.8635	&4.960	&0.2866 \\
            & \multicolumn{1}{l}{DID \cite{maharjan2019improving}}   & 7735.8G & 2.5M   &28.97	&0.7882	 &7.308  &0.1745  &31.58	&0.8524	&5.0702	&0.2826\\
            & \multicolumn{1}{l}{SGN \cite{gu2019self}}   &664.3G   &4.1M   &29.09	&0.7869	&7.050	&0.1773  &32.02	&0.8655	&4.598	&{\color{blue}\underline{0.2732}} \\
            & \multicolumn{1}{l}{SID \cite{chen2018learning}}   &560.1G & 7.7M   &29.10	 &0.7898  &7.081  &0.1721  &31.74	&{\color{blue}\underline{0.8709}}	 &4.836  &0.2736\\                     
            \hline
            \multirow{3}{*}{Two-Stage} & \multicolumn{1}{l}{LDC \cite{xu2020learning}}     &1699.4G    &8.6M    &29.93	&0.7927	&6.502	&0.1723  &30.46  & 0.8500   &5.400  &0.3131  \\
            & \multicolumn{1}{l}{MCR \cite{dong2022abandoning}}  &1046.1G     &15.0M    &30.16	&0.7949	&6.290	&{\color{blue}\underline{0.1683}}  &32.22	&0.8555	&4.534	&0.2839 \\
            & \multicolumn{1}{l}{DNF \cite{jin2023dnf}}   &658.4G   &2.8M    &{\color{blue}\underline{30.47}}	&{\color{blue}\underline{0.7971}}	&{\color{blue}\underline{6.189}}	&0.1728 &{\color{blue}\underline{33.18}}	&\textbf{{\color{red}0.8721}}	&{\color{blue}\underline{3.953}}	&0.2789\\
            \hline
            \multirow{1}{*}{\textbf{Single-Stage}} & \multicolumn{1}{l}{\textbf{Ours}}   &393.5G   &1.7M    &\textbf{{\color{red}30.72}}	&\textbf{{\color{red}0.7994}}	&\textbf{{\color{red}6.053}}	&\textbf{{\color{red}0.1681}}  &\textbf{{\color{red}33.49}}	&0.8690	&\textbf{{\color{red}3.812}}	&\textbf{{\color{red}0.2689}} \\
            \hline
    \end{tabular} }
    \label{sota}
\end{table*}

\begin{figure*}[t]
    \centering	   
    \centering{\includegraphics[width=17.35cm]{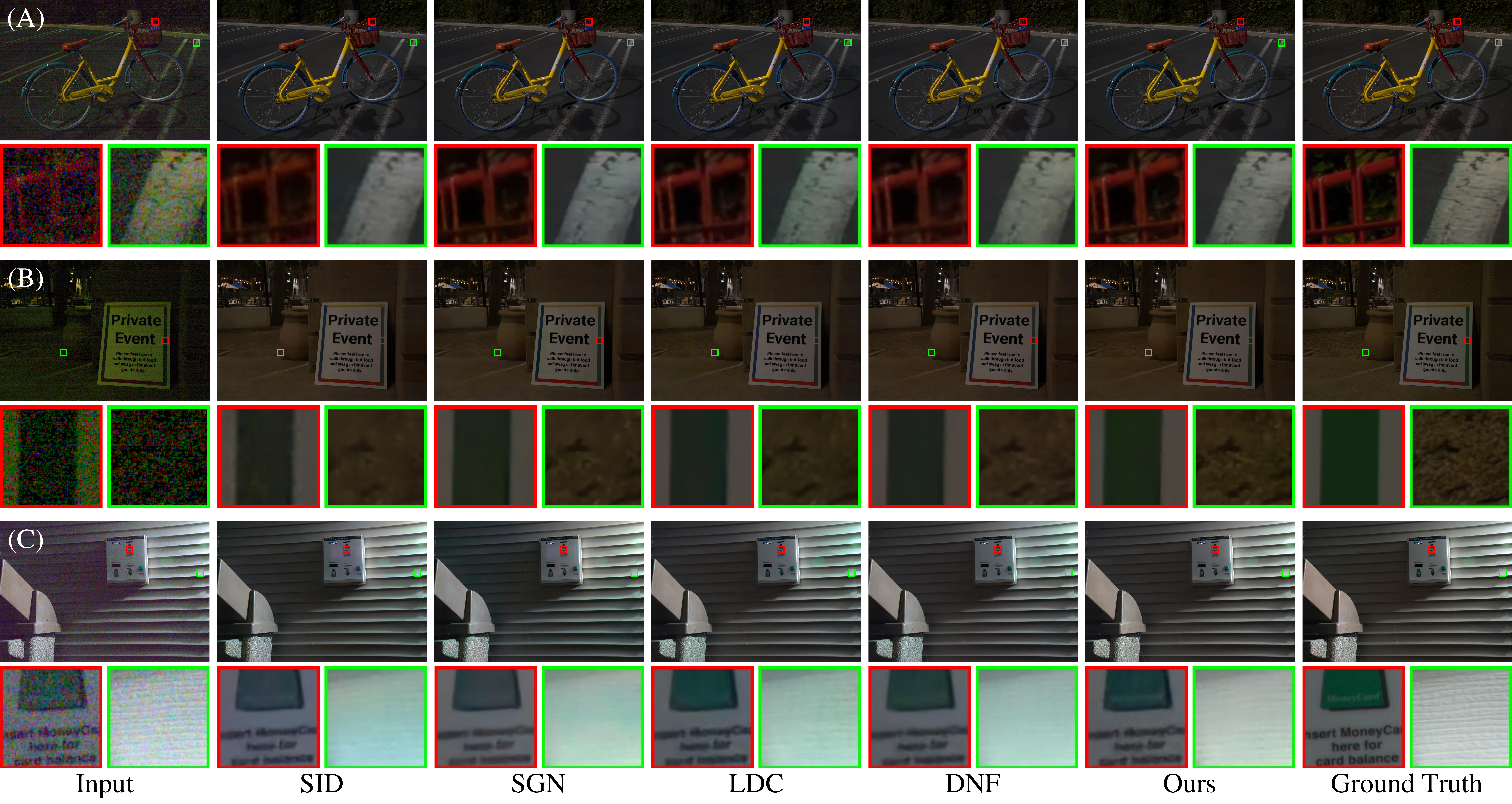}}  
    \caption {Visual comparisons between our FDANet and the state-of-the-art methods on the SID-Sony dataset (Zoom-in for best view). We amplified and post-processed the input images with an ISP for visualization.}
    \label{visual_sid}
\end{figure*}

\begin{table*}[t]
    \centering
    \caption{Quantitative results of raw-based LLIE methods on the MCR \cite{dong2022abandoning} and SID-Fuji \cite{chen2018learning} datasets. "-" indicates the result is not available.}
    \small
    \scalebox{1}{
        \begin{tabular}{clllllcccc}
            \hline  
            \multirow{2}{*}{Category} & \multirow{2}{*}{Method}  & \multicolumn{4}{c}{MCR\cite{dong2022abandoning}} & \multicolumn{4}{c}{Fuji\cite{chen2018learning}} \\  
            \cline{3-10}
            &     &  PSNR$\uparrow$  & SSIM$\uparrow$ & $\Delta E$$\downarrow$  & LPIPs$\downarrow$ &  PSNR$\uparrow$  & SSIM$\uparrow$ & $\Delta E$$\downarrow$  & LPIPs$\downarrow$  \\  
            
            \hline
            \multirow{5}{*}{Single-Stage} & 
            \multicolumn{1}{l}{LLPackNet \cite{Lamba_Balaji_Mitra_2020}}   &26.06	&0.8470	&6.062	&0.0896  & - &-    &-  &-\\
            & \multicolumn{1}{l}{IRT \cite{Lamba_Mitra_2021}}   & 26.76	&0.8404	&5.832	&0.0827  &27.02	&0.7100	&9.340	&0.2282\\
            & \multicolumn{1}{l}{DID \cite{maharjan2019improving}}   & 27.19	&0.8991	&5.684	&0.0498 & {-} &{-}    & - &-\\
            & \multicolumn{1}{l}{SGN \cite{gu2019self}}   &27.39	&0.8875	&5.610	&0.0556  &27.36	&0.7180	&9.313	&0.2187 \\   
            & \multicolumn{1}{l}{SID \cite{chen2018learning}}     &29.59	&0.9058	&4.108	&{\color{blue}\underline{0.0493}}  &27.30	&0.7193	&9.249	&0.2100\\     
                   
            \hline
            \multirow{3}{*}{Two-Stage} & \multicolumn{1}{l}{LDC \cite{xu2020learning}}     &27.12	&0.8767	&5.809	&0.0617  &28.26	&0.7220	&8.466	&{\color{blue}\underline{0.2086}}\\
            & \multicolumn{1}{l}{MCR \cite{dong2022abandoning}}   &31.28	&0.9065	&3.561	&0.0495  &-  &-    & - &-\\
            & \multicolumn{1}{l}{DNF \cite{jin2023dnf}}   &{\color{blue}\underline{31.47}}	&{\color{blue}\underline{0.9074}}	&{\color{blue}\underline{3.358}}	&0.0516  &{\color{blue}\underline{28.66}}	&{\color{blue}\underline{0.7250}}	&{\color{blue}\underline{8.189}}	&0.2087\\
            \hline
            \multirow{1}{*}{\textbf{Single-Stage}}  & \multicolumn{1}{l}{\textbf{Ours}}  &\textbf{{\color{red}31.79}}  &\textbf{{\color{red}0.9116}}    &\textbf{{\color{red}3.243}}  &\textbf{{\color{red}0.0492}} &\textbf{{\color{red}28.88}}	&\textbf{{\color{red}0.7278}}	&\textbf{{\color{red}8.016}}	&\textbf{{\color{red}0.2075}}    \\
            \hline
    \end{tabular} }
    \label{mcr_fuji}
\end{table*}

\begin{figure*}[htp]
    \centering	   
    \centering{\includegraphics[width=17.35cm]{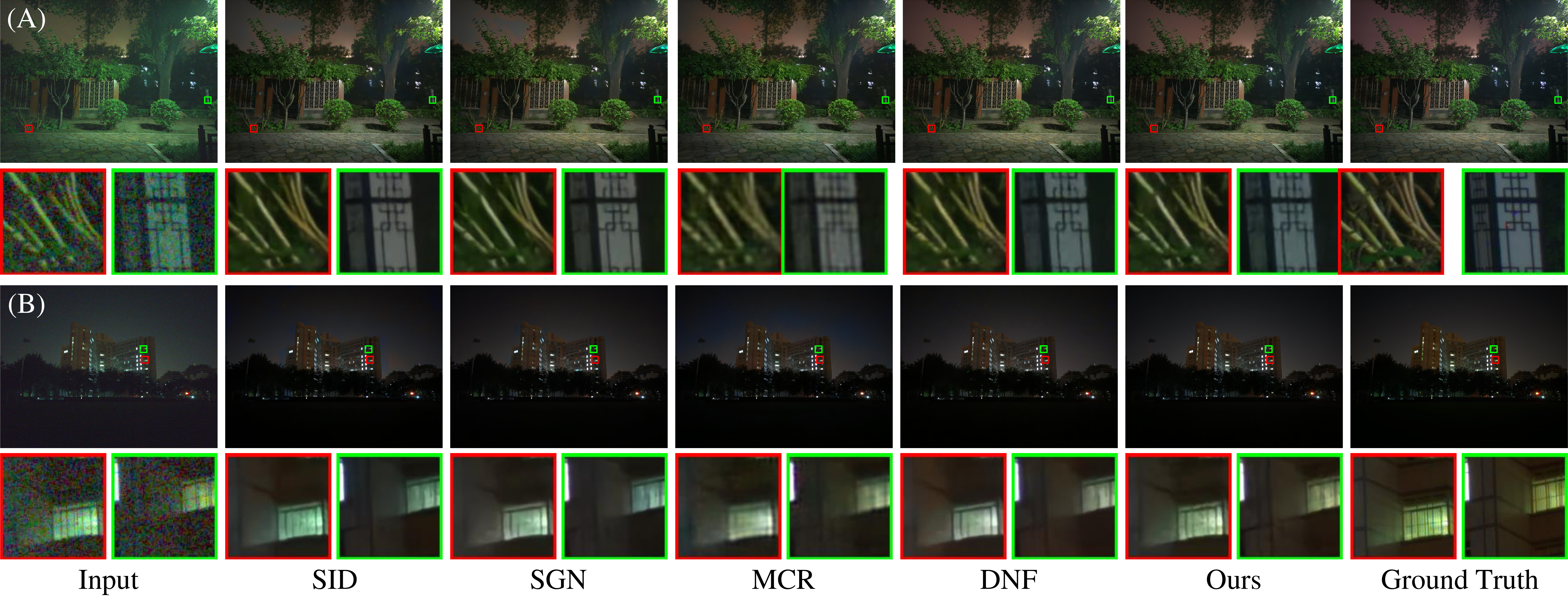}}  
    \caption {Visual comparisons between our FDANet and the state-of-the-art methods on the LRID dataset (Zoom-in for best view).}
    \label{visual_imx}
\end{figure*}

\subsection{Color Mapping Branch}
The color mapping branch is constructed by a raw encoder $E_{\text{raw}}$, a feature domain adaptation module, and a sRGB decoder $D_{\text{rgb}}$. In order to explore local correlations with small computation costs, we utilize the channel-independent denoising (CID) blocks proposed in \cite{jin2023dnf} to construct the raw encoder. Each block consists of a depth-wise convolution with a $7 \times 7$ kernel, two point-wise convolutions with a $1 \times 1$ kernel, and a GELU function. 

\textbf{Feature domain adaptation.} It is difficult to make the encoded features $F_{\text{en}}$ to fit both the denoising and color mapping tasks. To address this issue, we propose the Feature Domain Adaptation (FDA) module, as shown in Fig. \ref{model}. The FDA module is designed to map the encoded raw features ($F_{\text{en}}$) to the sRGB domain $F^{'}_{\text{en}}$. Since $F_{\text{en}}$ is also constrained by the raw denoising task, its features are usually smooth, and the global tone belongs to the raw feature domain. Therefore, the FDA module should change the global brightness and color tone and refine the local details of $F_{\text{en}}$. On the other hand, the practical imaging process is usually line-based in the image sensor. To improve the efficiency of the inference process, the FDA should have low complexity and fewer line buffers.  

Based on these observations, we propose a Line-Based Transformer to serve as the key part of the FDA module. Different from the global transformer, which explores global self-attention \cite{dosovitskiy2020image, Zamir_Arora_2022}, and local transformer, which explores local self-attention inside a window \cite{hassani2023neighborhood, chen2023comparative}, we calculate global attention along the line (width) dimension, and calculate local attention along the column (height) dimension. In other words, for a query token $Q_i$, the key tokens are generated from a $h\times W$ region, where $W$ is the width of the image, and $h$ is the local height centered at $Q_i$.  For query tokens in the same line, they share the same key tokens. 
Specifically, given an encoded feature $F_{\text{en}} \in \mathbb{R}^{H\times W\times C}$, we utilize LayerNorm, point-wise convolution (PConv), and depth-wise convolution (DConv) layers to generate the query query ($Q\in \mathbb{R}^{H\times W\times C}$), key ($K\in \mathbb{R}^{H\times W\times C}$), and value ($V\in \mathbb{R}^{H\times W\times C}$).
Then, for a query $Q_i\in \mathbb{R}^{1\times C}$, we localize our attention to the region with $h$ lines. The transformed feature $F_i\in \mathbb{R}^{1\times C}$ for $Q_i$ is generated by 
\begin{equation}
  F^t_i = Q_i \sum\limits_{j=1}^{hW}({K_j^TV_j})=Q_i({K_{\Omega{(i)}}^TV_{\Omega{(i)}}}),
  \label{eq:vit}
\end{equation}
where $K_{\Omega{(i)}}$ represents the keys in a $hW$ region centered at $Q_\text{i}$ along the height dimension. Note that, we did not utilize the traditional transformer with softmax operations. We remove the softmax function following \cite{katharopoulos2020transformers, tsai2019transformer} to accelerate the attention calculating process. After calculating the attention for all the query tokens, we obtain the transformed feature $F^\text{t} \in {\mathbb{R}^{H\times W\times C}}$. Hereafter, we employ Group Normalization to stabilize the training process and enhance the feature transformation quality by balancing the output variance. Finally, we use PConv and DConv to refine the local details further.

In this way, our Lineformer only has a complexity of $O(n)$ since $h$ is much smaller than $H$, which is much faster than the ViT \cite{dosovitskiy2020image} with a complexity of $O(n^2)$. Our design satisfies the requirement of line operations in hardware and significantly reduces the line buffer occupancy in the hardware implementation. In addition, since the attention is calculated across the whole line, the global color tone along the line direction can also be well balanced. Note that, our Lineformer is applied on multi-scale features of the single encoder, which can avoid information loss.

In the decoding stage, the adapted features $F^{'}_{\text{en}}$ are fed into the sRGB decoder $D_{\text{rgb}}$ to generate the final output $\hat{Y}_{\text{rgb}}$ in the sRGB domain. To ensure global color consistency, the sRGB decoder is implemented using a Global Transformer \cite{Zamir_Arora2022, jin2023dnf}.

\subsection{Raw Denoising Branch}
If we only utilize the color mapping branch and supervise the final result with sRGB domain ground truth $Y_{\text{rgb}}$, the network will pay more attention to the cross-domain mapping task other than the denoising task since the loss introduced by the misaligned mapping is larger than the residual noise. To alleviate this problem, we further introduce an auxiliary raw denoising network as an intermediate supervisor. The raw denoising network shares the same encoder with the color mapping network but has a separate raw decoder. Compared with the other two-stage raw enhancement networks, our method reduces redundant encoders and avoids information loss. 
Similar to the encoder, the raw decoder is also composed by multiple CID modules.  

In the training phase, we use clean raw and clean sRGB as supervision to accomplish both the denoising and color mapping tasks. The combined loss function is
\begin{equation}
  L = \Vert{Y}_{\text{rgb}} - \hat{Y}_{\text{rgb}}\Vert_1 + \Vert{Y}_{\text{raw}} - \hat{Y}_{\text{raw}}\Vert_1,
  \label{eq:loss}
\end{equation}
where $\hat{Y}_{\text{raw}}$ and $\hat{Y}_{\text{rgb}}$ are the results of the color mapping branch and denoising branch, respectively. Note that, during inference, the auxiliary raw supervision network is deactivated. Thus, our approach is a single-stage framework, as shown by the orange arrows in Fig. \ref{model}, which significantly reduces the computational cost compared with the two-stage methods.

\section{Experiments}
\label{sec:exp}

\subsection{Experimental Settings}
\textbf{Datasets.} We evaluate the performance of our proposed FDANet on four datasets that include raw-sRGB pairs. They represent various scenes and are captured by different devices. These datasets include SID, captured by DSLR cameras (SONY \& Fuji) \cite{chen2018learning}, LRID, captured by smartphone cameras (Redmi K30) \cite{feng2023learnability}, and MCR, captured by security cameras (Onsemi) \cite{dong2022abandoning}. We follow previous works and exclude images in the SID dataset with noticeable misalignments. The LRID dataset is used for raw denoising, which we have reorganized to fit the requirements of our task. For the MCR dataset, we have selected the image with the longest exposure time for each scene as the raw ground truth since there is no raw format ground truth available. In the Supplementary Material, we evaluate the SID dataset with dark shading correction and the MCR dataset with monochrome raw data as intermediate supervision. \emph{For a fair comparison, we retrained and validated all comparison methods on the same training and testing sets.}\\
\textbf{Metrics.} We evaluate model performance from four aspects: pixel level, structural, perceptual quality, and color fidelity. Specifically, we report traditional PSNR and SSIM metrics on the RGB channel to evaluate the reconstruction accuracy. We also employ LPIPS \cite{zhang2018unreasonable} to evaluate image perceptual quality. In addition, we use the CIELAB color space \cite{zhang1996spatial, hill1997comparative, Liu_He_Chen_Zhang_Zhao_Dong_Qiao_2021} (also known as $\Delta E$) to evaluate the difference in color dimensions of the reconstructed images.\\
\textbf{Implementation Details.} We implement our model with Pytorch on the RTX 3090 GPU platform. We have set the batch size to 1. For optimization, we use AdamW optimizer \cite{Xiao_Singh_Mintun_Darrell_Dollár_Girshick_2021} with $\beta_{1}=0.9$, $\beta_{2}=0.99$. The learning rate is initialized to $2\times 10^{-4}$.

\begin{figure*}
    \centering	   
    \centering{\includegraphics[width=16cm]{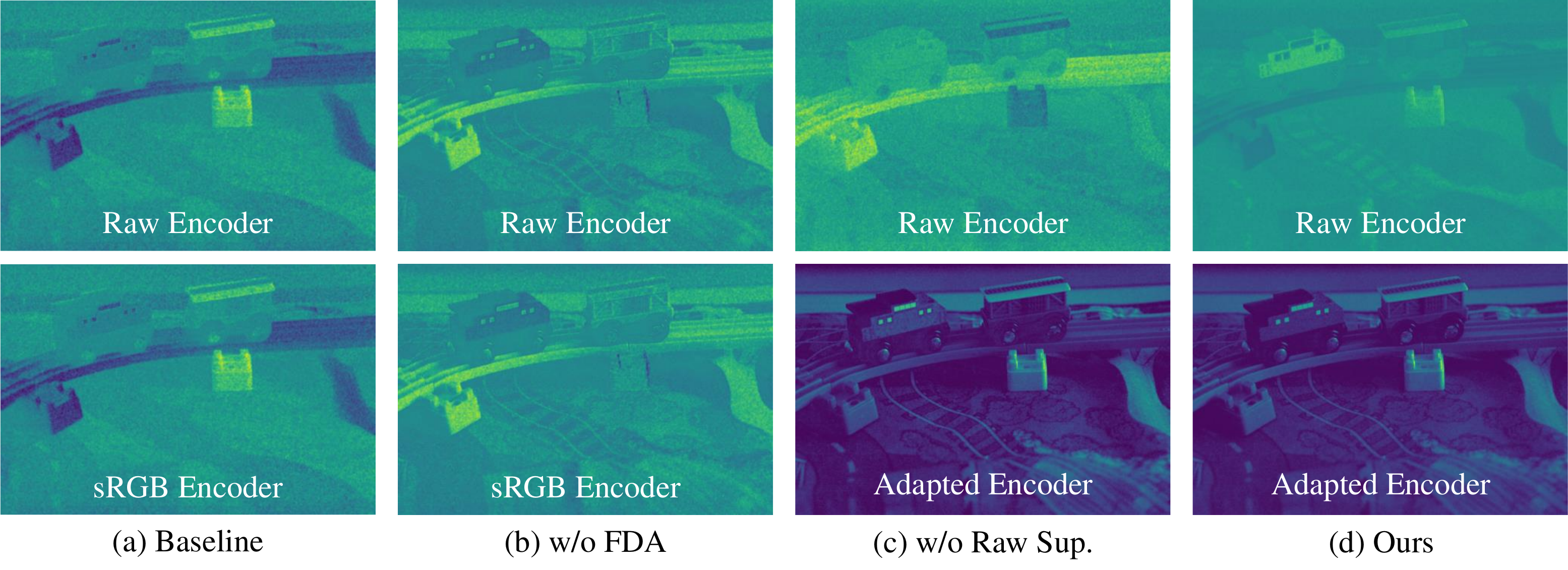}}  
    \caption {Feature visualization for ablation study. The features are extracted by the corresponding encoder. The raw encoder means the input of the encoder is original raw noisy images. The sRGB encoder means the output of the encoder is directly fed into a sRGB decoder. Therefore, the raw and sRGB encoders are the same in (a), (b). } 
    \label{fea_ablation}
\end{figure*}

\subsection{Comparison with State-of-the-Arts}
\textbf{Quantitative comparison.} We quantitatively compare the proposed method with a wide range of state-of-the-art raw-based LLIE methods in Tab. \ref{sota} and Tab. \ref{mcr_fuji}, including the single-stage methods, SID \cite{chen2018learning}, DID \cite{maharjan2019improving}, SGN \cite{gu2019self}, LLPackNet \cite{Lamba_Balaji_Mitra_2020}, and RRT \cite{Lamba_Mitra_2021}, as well as the two-stage methods, LDC \cite{xu2020learning}, MCR \cite{dong2022abandoning}, and DNF \cite{jin2023dnf}. Our FDANet outperforms SOTA methods on four datasets with fewer computational and memory costs. Compared with the recent best two-stage method DNF (CVPR 2023) \cite{jin2023dnf}, FDANet achieves a notable improvement of 0.25, 0.31, 0.32, and 0.22 dB on SID-Sony, LRID, MCR, and SID-Fuji. Moreover, our method costs only $60.7\%$ (1.7 / 2.8) Parmas and $59.7\%$ (393 / 658) FLOPs compared to DNF. Additionally, our method demonstrates superiority in terms of fidelity, perceptual, and color metrics, as shown in Tab. \ref{sota} and Tab. \ref{mcr_fuji}. Notably, on the SID dataset, the $\Delta E$ metric has been reduced by 0.12.\\
\textbf{Qualitative Results.} The visual comparisons of FDANet and state-of-the-art raw-based LLIE methods are shown in Fig. \ref{visual_sid} and Fig. \ref{visual_imx}. Please zoom in for a better view. Previous methods either result in color distortion (e.g., SID, LDC, DNF in Fig. \ref{visual_sid} (B) and (C), SGN, MCR in Fig. \ref{visual_imx} (A)), or fail to suppress noise (e.g., LDC, DNF in Fig. \ref{visual_sid} (B), SID, MCR in Fig. \ref{visual_imx} (B)). In some results, the texture details are lost (e.g., MCR, DNF in Fig. \ref{visual_sid} (A) and (C). Benefiting from the Feature Domain Adaptation module with a Lineformer and raw supervision, our FDANet successfully suppresses noise while achieving vivid texture and accurate color reconstruction.
Our method effectively reduces significant noise while retaining intricate texture details. Additionally, our method ensures precise color reconstruction and vivid color saturation. More visual comparisons, as well as full-resolution comparisons, are presented in our supplementary material.

\begin{table}[t]
\centering
    \caption{Ablation study on the FDA and raw supervision, denoted as Raw Sup.}
    \begin{tabular}{cccccc}
        \hline
        {Variants} & FDA         &{Raw Sup.}  & PSNR$\uparrow$ & SSIM$\uparrow$ & $\Delta E$$\downarrow$ \\
        \hline
        \#1     & \XSolidBrush  & \XSolidBrush   & 30.02	     & 0.7917	& 6.320           \\
        \#2     & \XSolidBrush  & \Checkmark     & 30.12  	     & 0.7941	 & 6.385           \\
        \#3     & \Checkmark   &  \XSolidBrush   & 30.29        & 0.7952    &    6.274         \\
        \#4     & \Checkmark   & \Checkmark      & \textbf{30.72}	    & \textbf{0.7994}   & \textbf{6.053}	\\
        \hline
    \end{tabular} 
    \label{factor}
\end{table}

\subsection{Ablation Studies}
\textbf{FDA and Raw Supervision.} 
To validate the effectiveness of the Feature Domain Adaptation module and raw supervision, we remove the two modules one by one. The baseline network means removing both two modules. Tab. \ref{factor} presents the results of different configurations. After removing the FDA, the performance is degraded by 0.6 dB due to the feature domain misalignment. Comparison between Variants \#1 and \#2 demonstrates that, despite having raw supervision, the model only shows minor improvement due to unsolved feature domain misalignment. If w/o Raw Sup. (\#3), the denoising task cannot directly constrain the encoded features. Therefore, the domain gap between raw encoded features and color mapping features is also a hybrid task, including noise removal and color mapping, which is difficult to solve. Therefore, variant \#3 is inferior to the full model by 0.43 dB.  



Additionally, we visualize the features of the four variants. As shown in Fig. \ref{fea_ablation}, the raw-encoded features suffer from severe feature ambiguity and noise when the FDA and raw supervision are removed. For the baseline and w/o FDA variants, the color tone of the sRGB encoder features is unsuitable for the sRGB decoder due to the lack of the FDA module.
For the variant w/o raw supervision, the features of the raw encoder are noisy. After feature adaptation, the color tone of adapted features changes but still contains noise.
By incorporating FDA and raw supervision, we can effectively solve feature domain misalignment problems and reduce noise.

\begin{table}
\centering
    \caption{Ablation on the Lineformer in FDA by replacing it with other modules.}
    \begin{tabular}{ccccc}
        \hline
        FDA        & PSNR$\uparrow$ & SSIM$\uparrow$ & $\Delta E$$\downarrow$ & FLOPs\\
        \hline
       Convolution \cite{long2015fully}         & 30.23 	        & 0.7941	        &   6.135    &  79.65G       \\
       SSA \cite{chen2023comparative}    & 30.32  	        & 0.7962	        &   6.224     &  52.27G      \\
       MCC \cite{jin2023dnf} & 30.41            & 0.7962            &   6.091     &  \textbf{33.99G}      \\
       \textbf{Ours (Lineformer)}                & \textbf{30.72}	    & \textbf{0.7994}   & \textbf{6.053}	& 34.04G  \\
        \hline
    \end{tabular} 
    \label{exe:ltd}
\end{table}

\textbf{Exploration of FDA Structure.} 
The Feature Domain Adaptation module is essential to transform raw-encoded features for sRGB color mapping. Our Lineformer is an efficient and effective implementation for the FDA module.  
To verify its effectiveness, we replace it with three different approaches: the spatial self-attention (SSA) module from X-Restormer \cite{chen2023comparative}, which is a local transformer with overlapping cross-attention; the matrixed color correction (MCC) from DNF \cite{jin2023dnf}, which is a finely-designed global attention module; and Fully Convolution \cite{long2015fully}. As demonstrated in Tab. \ref{exe:ltd}, our Lineformer outperforms the MCC module by 0.31 dB, but with similar FLOPs. The convolution-based and local transformer based solutions are also inferior to our solution. In summary, our Lineformer, which explores local and global correlations, is a good solution for the FDA task in raw LLIE. 


\section{Conclusion}
\label{sec:conclusion}
We propose a single-stage network to decouple the denoising and color mapping tasks in raw LLIE, which have both the advantage of a two-stage enhancement process and the efficiency of single-stage inference. Our Lineformer is an efficient FDA module, which 
bridges the denoising features and color mapping features with less computation cost and line buffers. Extensive results on four datasets demonstrate the superiority of our method. Our FDA strategy can also be applied to other raw image based restoration tasks, and we expect more works along this avenue.



{
    \small
    \bibliographystyle{ieeenat_fullname}
    \bibliography{main}
}

\end{document}